\documentclass{midl} % Include author names
\usepackage{amsmath}

\usepackage{graphicx}

\usepackage{caption}

\jmlrproceedings{MIDL 2020}{Medical Imaging with Deep Learning 2020}
\jmlrvolume{}
\jmlryear{}
\jmlrworkshop{MIDL 2020 -- Short Paper}
\editors{}

\usepackage{mwe} % to get dummy images
\usepackage{amssymb}
\title[Functional Space Variation Inference]{Functional Space Variational Inference for Uncertainty Estimation in Computer Aided Diagnosis}

\midlauthor{\Name{Pranav Poduval{}\nametag{}} \Email{pranav97.poduval@gmail.com}\\
\Name{Hrushikesh Loya\nametag{}} \Email{loyahrushikesh@gmail.com}\\
\Name{Amit Sethi \nametag{}} \Email{amitsethi@gmail.com}\\
\addr  Department of Electrical Engineering, Indian Institute of Technology, Bombay \AND
}

\begin{document}

\maketitle

\begin{abstract}
Deep neural networks have revolutionized medical image analysis and disease diagnosis. Despite their impressive performance, it is difficult to generate well-calibrated probabilistic outputs for such networks, which makes them uninterpretable black boxes. Bayesian neural networks provide a principled approach for modelling uncertainty and increasing patient safety, but they have a large computational overhead and provide limited improvement in calibration. In this work, by taking skin lesion classification as an example task, we show that by shifting Bayesian inference to the functional space we can craft meaningful priors that give better calibrated uncertainty estimates at a much lower computational cost.
% Despite their impressive performance, they have been criticized for being non-interpretable black box. Standard approach for uncertainty estimation in neural networks is by approaching problem from a Bayesian view, by placing a prior on the parameters and approximating the posterior. In this work we show that by shifting to the functional space we can craft more meaningful prior's https://www.overleaf.com/project/5e21a98a1d93f30001601089which in turn give us far better calibrated uncertainty estimates at a significantly lower computational cost.
\end{abstract}


\section{Introduction}

In computer-aided diagnosis, AI models must not only be accurate, but they should also indicate when they are likely to be incorrect. For instance, control should be passed on to human doctors when the confidence of a neural network for disease diagnosis is low. Model calibration is the degree to which a model’s predicted probabilities reflect the true correctness likelihood. Calibrated confidence estimates are also important for model interpretability and they provide a valuable extra bit of information to establish trustworthiness with the user. This is important for deep neural networks, whose classification decisions are often and difficult to interpret.

It is well known that popular neural network frameworks only provide a point estimate of the true underlying distribution. Furthermore, the typical classification setting of training the softmax output layer using cross-entropy loss typically gives “over-confident” (low entropy) class probability mass distributions, even when there is a classification error. This is especially concerning for training on medical datasets that are often relatively smaller and suffer from severe class imbalance \citep{esteva2017dermatologist}. In other words, the popular deep learning models give poorly calibrated uncertainty estimates for cases that are ambiguous, or difficult, or  \textit{out-of-distribution} (OOD), including those from a new class.

% Recently deep learning methods have received significant attention for computer aided diagnosis in a variety of medical imaging domains \citet{esteva2017dermatologist}. They outperform former methods on performance metrics such as accuracy, F1 score, or ROC-AUC etc. However care has to be taken before deploying deep neural networks to support medical diagnosis. This is because it is known that neural networks are only a point estimate of the true underlying distribution, and the softmax output layer that is used to get a probability score is typically “over-confident” for one class. This issue causes most deep learning models to be poorly calibrated and give over-confident estimates for ambiguous or unknown cases.

Bayesian modelling offers a set of tools to reason about uncertainty. Existing Bayesian approaches involve approximate inference using either Markov Chain Monte Carlo \citep{neal2011mcmc} or variational inference methods, such as dropout \citep{gal2016dropout}. This idea has attracted attention of the medical community to ensure that difficult cases for computer-aided diagnosis are duly flagged for review \citep{laves2019uncertainty}. Since most Bayesian neural networks (BNNs) have their prior defined on the weight space, the regularization caused by these prior is not able to calibrate the network output, nor do these priors explicitly make the model under-confident on the OOD samples. We show that by performing variational inference on the functional space we can craft a prior that is able to simultaneously calibrate the network as well as ensure the the recognition of OOD samples as more uncertain. Our method is also significantly less computationally expensive as compared to Bayesian or frequentist approaches. Although our method shares some similarities with Evidential Deep Learning (EDL) \citep{sensoy2018evidential}, it has been derived from a variational Bayesian framework and it can distinguish distributional versus data uncertainties (shift in distribution versus class confusion, respectively), unlike EDL.

\section{Proposed Method}

For classification among $K$ classes, deep neural networks represent a function $f_{\theta}:X \rightarrow \textbf{p} \in [0,1]^K$, where $X$ represents the input, and $\textbf{p}$ represents a probability mass function such that $\sum_{i=1}^Kp_i = 1$. The output distribution $p(Y|X,\theta) = \text{Cat}(Y|\textbf{p})$. A prior on $\theta$ implicitly defines a prior measure on the space of $f(X)$, denoted as $p(f)$. Priors of convenience on $\theta$, such as a fully factorized Gaussian, are often used, and it is difficult to formulate a prior on the weight space that is informative in the sense that it leads to high uncertainty on OOD examples. We therefore define a uniform prior on the $K$-dimensional unit simplex for the functional space, such that $p(f) = D(\textbf{p}|\langle1,\dots,1\rangle)$ (completely uncertain prior). While it seems intuitively satisfying to have a model that is not biased towards ``over-confident" outputs (towards which the usual cross-entropy loss is severely biased), we also empirically show that such a uniform prior gives well-calibrated outputs. %Completely uncertain prior is added

Given the training data $D = (X^D,y^D)$ and the test points $(x^*,y^*)$ we have:
\begin{equation}
p(y^*|x^*,D) =  \int p(y^*|\textbf{p})\, p(\textbf{p}|x^*,D)  \,d\textbf{p} 
\end{equation}
We assume $p(y^*|\textbf{p}) = \text{Cat}(y^*|\textbf{p})$. We further assume that the neural network estimates a Dirichlet distribution $\text{Dir}(\textbf{p}|\alpha)$ with $\alpha>0$, as done by \citet{sensoy2018evidential}, because of its analytical tractability.  
In other words, unlike for a standard neural network where $\textbf{p} = f_{\theta}(x)$ is the point estimate output, in our case $\text{Dir}(\textbf{p}|\alpha) = q_{\theta}(f(x))$ is the marginal functional distribution. This is similar to how a Gaussian process has a multivariate Gaussian as its marginal distribution.

The true functional posterior $p(f|D)$ is intractable, but it can be approximated by minimizing the functional evidence lower bound (fELBO) as done by \citep{sun2019functional}:
%such that the objective is:

\begin{equation}
\label{eq:felbo}
\mathcal{L}(q) = -\mathbb{E}_{ q(f)}[\log p(y^D|f(X^D))] + \text{KL}[q(f)||p(f)]
\end{equation}

The second term in Equation~\ref{eq:felbo} is the functional KL divergence, which is hard to estimate. Therefore, we shift to a more familiar metric, the KL divergence between the marginal distributions of function values at finite sets of points  $\textbf{x}_{1:n}$. \citep{sun2019functional} has shown:
\begin{equation}
    \text{KL}(q(f)||p(f))=\smash{\displaystyle\sup_{\textbf{x}_{1:n} }} \mathop{\mathbb{\text{KL}}}\left[ q(f(\textbf{x}_{1:n})||p(f(\textbf{x}_{1:n}) \right]  = \sum_{i=1}^n \text{KL}[\text{Dir}(\textbf{p}|\alpha_i)||\text{Dir}(\textbf{p}|\langle 1, \dots 1 \rangle)]
\end{equation}
A more relaxed way of sampling these “measure points" $\textbf{x}_{1:n}$, is to assume $\textbf{x}_{1:k} \sim X^D$ (training distribution) and $\textbf{x}_{k+1:n} \sim c$ where $c$ is a distribution having the same support as the training distribution, which could be OOD samples, that can be forced to be more uncertain. This approach is similar to \cite{hafner2018noise}, \cite{malinin2018predictive}.
%  Note $y$ is assumed to be a one-hot vector, such that if the true label is $j$ then $y_j =1$ and $y_i = 0$ for all $i \neq j$. So now we can get the closed form solution of the first term in loss function as -

We get a closed form solution for the first part in Equation~\ref{eq:felbo} by assuming $y$ to be a one-hot vector as follows:
\begin{equation}
\mathcal{L}_1 = \int \left[\sum_{i=1}^K -y_i\log p_i \right] \frac{1}{B(\alpha)}\prod_{i=1}^K p_i^{\alpha_i-1} d\textbf{p}  = \sum_{i=1}^K y_i \left( \digamma(\sum_{j=1}^K \alpha_j)-\digamma(\alpha_i) \right)
\end{equation}
% where $\digamma(.) $ is the digamma function. Thus, both parts of Equation~\ref{eq:felbo} have analytically closed forms, which helps in smooth training as we do not have to rely on the noisy Monte Carlo estimates, unlike previously proposed Bayesian neural networks.

 $\digamma(.)$ is the digamma function. To measure calibration of the proposed model we group predictions $p \in [0,1]$ into $M$ bins each of size $\frac{1}{M}$, and let $B_m$  be the set of indices of samples whose prediction confidence falls into the interval $(\frac{m-1}{M},\frac{m}{M}]$ for $m \in \{1, \dots, M\}$. Now we define accuracy of $B_m$ as acc($B_m$)= $\frac{1}{|B_m|}\sum_{i \in B_M} 1_{\{\hat{y}_i = y_i\}}$, where $\hat{y}$ is the predicted outcome with confidence $p$, and $y$ is the true label. Similarly, we define the average confidence as conf($B_m$) = $\frac{1}{|B_m|}\sum_{i \in B_m} p_i$. For perfect calibration we expect acc($B_m$) = conf($B_m$). In order to quantify how well calibrated our networks are, we use Expected Calibration Error (ECE) $= \sum_{i=1}^M \frac{|B_m|}{n}|\text{acc}(B_m) - \text{conf}(B_m)|$ as the metric. Note ECE = 0 for perfect calibration.

\section{Results}
% We apply our method to the problem of skin lesion classiﬁcation, using the HAM10000 dataset \citet{tschandl2018ham10000}. This dataset contains 10,015  images of common pigmented skin lesions, divided over seven classes. Similar to other medical datasets, this dataset is heavily imbalanced. The images are down-scaled to 224×224 pixels, and normalized. We randomly split the full dataset into a training set (9,013 images), a validation set, and a hold-out test set (both 501 images). We use the training and validation set to train a deep neural network architecture and optimize hyperparameters. For training we use the ResNet18 \citet{he2016deep} architecture and the trainable parameters are optimized using the Adam, with an initial learning rate of 0.0001. 
% \par
We applied our method to the problem of skin lesion classification, using the HAM10000 dataset \citep{tschandl2018ham10000}. ResNet 50 architecture optimized by Adam was used.
% We trained a ResNet50 architecture using the Adam optimizer, with an initial learning rate of 0.0001.

% For example, if we have a well calibrated weather prediction model that predicts sunny event with 80\% probability for 100 days then, any  deviation from 80 sunny days and 20 non-sunny days will imply a poorly calibrated model. A poorly calibrated model is hard to interpret and is too unreliable to be deployed in the real world. This is especially true when a classification model is not trained for its own sake but instead for the purpose of passing on such probabilities to some other decision-making component, which is often the case in medical domain. 

From Table 1 we can see that although standard Bayesian approaches do help calibrate the model, our method has a significantly lower ECE. That too at a much lower computational cost, approximately 25x less computationally expensive than Dropouts (monte carlo approximation) and 5x more memory efficient than Ensembles (ensemble size). 
%, with only a marginal compromise in test accuracy.

% Because minimizing a cross entropy loss does not ensure calibration, and even tends to over-fit classification accuracy \citet{guo2017calibration}, it’s imperative to calibrate any model where probabilities are passed on to some other decision making system, which is often the case in practical medical applications. Regular Bayesian Neural Networks like Bayes by Backprop (BBB) \citet{blundell2015weight}, Dropouts \citet{gal2016dropout} etc. tend to be better calibrated than the standard Neural Networks but are far from satisfactory, since the regular prior used in these cases are not able to prevent under-estimation or over-estimation of class probabilities.

% \begin{table}[!htb]
%   \caption{Comparison of classification accuracy and ECE on HAM10000 dataset for the proposed and the other Bayesian approaches}
%   \label{accuracytable}
%   \centering
%   \begin{tabular}{|l|l|l|}
%     \toprule
%       \hline         
%     Method & Test Accuracy & ECE\\
%     \midrule
% \hline
%     Standard NN  & 84.38\% & 7.73\% \\
%     \hline
%     Concrete Dropout & \textbf{86.82\%} & 6.39\%\\
%     \hline
%     Deep Ensemble & 85.21\% & 3.12\%\\
%     \hline
%     Bayes-By-Backprop & 84.29\% & 7.70\%\\
%     \hline
%     Functional Space VI & 84.84\%& \textbf{1.17}\%\\
%     \hline
%     \bottomrule 
%   \end{tabular}
% \end{table}

\begin{table}[h!]
  \caption{Comparison of classification accuracy and ECE on HAM10000 dataset }
  \label{accuracytable}
  \centering
  \begin{tabular}{|l|l|l|l|l|l|}
    % \toprule
       \hline         
    Method & Standard NN & Dropout & Deep Ensemble & Functional Space VI \\
    % \midrule
\hline
    Test Accuracy & 84.38\% & \textbf{86.32}\% &  85.21\%  &84.84\%\\
    \hline
    ECE (M = 15) & 7.73\% & 6.39\% &3.12\% & \textbf{1.17}\%\\
    % \hline
    % Deep Ensemble & 85.21\% & 3.12\%\\
    % \hline
    % Bayes-By-Backprop & 84.29\% & 7.70\%\\
    % \hline
    % Functional Space VI & 84.84\%& \textbf{1.17}\%\\
  \hline
    % \bottomrule 
  \end{tabular}
\end{table}

\par
% \textbf{Experiment to show benefits of uncertainty yet to think of
% }
% Refer to the Appendix for our work on quantifying different types of uncertainty.
% In the Appendix we show how our model can distinguish distributional uncertainty from data uncertainty, allowing it distinguish OOD samples much better than the previous Bayesian approaches, while also detecting samples within the distribution that are overlapping.
The entropy $\mathcal{H}[p(y|x,D)]$ is a measure of total uncertainty, whereas differential entropy $\mathcal{H}[D(\textbf{p}|\alpha)] $ is a measure of the distributional uncertainty.(See Appendix A for more details)%, both of which have an analytically closed form.
\section{Conclusions}
% Our method shares similarity with Evidential Deep Learning (EDL) \citet{sensoy2018evidential}, but has been derived completely from a Variational Bayesian Framework, also unlike EDL our model can distinguish Distributional Uncertainty from Data Uncertainty. 
We proposed a novel Bayesian NN framework whose prior explicitly forces OOD samples to become unconfident as well as allow us to estimate uncertainty analytically at test time, without needing approximate or expensive algorithms. We have also shown that our model gives well-calibrated uncertainty outputs, which can increase patient safety and assist a transfer of AI systems into clinical settings by including trustworthiness as a design factor in machine learning models for medical diagnosis. Our method is also significantly more computationally efficient making it a more viable option for resource-constrained problems.

\bibliography{midl-shortpaper.bib}

\appendix
\newpage
\section{Quantifying Uncertainty}

We use two measures to estimate uncertainty -- differential entropy and output entropy. The output entropy is a measure of the total uncertainty, where as the differential entropy is a good measure of distributional uncertainty. Output entropy is high whenever we encounter overlap between classes or we encounter samples from OOD. On the other hand, the differential entropy is high only when we encounter OOD samples and remains low even in case of data uncertainty \citep{malinin2018predictive}. 

Output entropy is defined as:
\begin{equation}\label{eq:example}
\mathcal{H}[p(y|x,D)] = -\sum_{i=1}^K p(y_i|x,D)\log(p(y_i|x,D))
\end{equation}
where
\begin{equation}\label{eq:example}
p(y_i|x,D) = \int p(y_i|\textbf{p}) \text{Dir}(\textbf{p}|\alpha) d\textbf{p} = \frac{\alpha_i}{\sum_{j=1}^K \alpha_j}
\end{equation}

Differential entropy is maximized when all categorical distributions are equiprobable. i.e. when posterior $q_{\theta}(f(x)) = D(\textbf{p}|\langle1,\dots,1\rangle)$, and it is defined as:
\begin{equation}\label{eq:example}
\mathcal{H}[D(\textbf{p}|\alpha)] = \log B(\alpha) + (\sum_{i=1}^K\alpha_i - K)\digamma(\sum_{i=1}^K\alpha_i)  - \sum_{i=1}^K(\alpha_i - 1)\digamma(\alpha_i) 
\end{equation}

% \begin{figure}
%     \begin{subfigure}[b]{0.25\textwidth}
%         \centering
%         \includegraphics[width=4cm]{20a.PNG} 
%         \caption{High Data Uncertainty}
%         \label{fig:gull}
%     \end{subfigure}%
%     \begin{subfigure}[b]{0.25\textwidth}
%         \includegraphics[width=4cm]{22a.PNG}
%         \caption{High Distributional Uncertainty}
%         \label{fig:gull2}
%     \end{subfigure}%
%     \caption{Fig (a) implies high data uncertainty so we will have low differential entropy \\Fig (b) has high distributional uncertainty so both uncertainty metrics will be high}
% \end{figure}

% \begin{figure}[h!]%
%     \centering
%     \subfloat[High Data Uncertainty]{\includegraphics[width=5cm]{20a.PNG} }%
%     \qquad
%     \subfloat[High Distributional Uncertainty]{\includegraphics[width=5cm]{22a.PNG} }%
%     \caption{Fig (a) implies high data uncertainty so we will have low differential entropy \\Fig (b) has high distributional uncertainty so both uncertainty metrics will be high}%
%     \label{fig:example}%
% \end{figure}

\begin{figure}[htbp]
\floatconts
  {fig:AppendixA}
  {\caption{(a) implies high data uncertainty so we will have low differential entropy \\(b) has high distributional uncertainty so both uncertainty metrics will be high}}
  {%
    \subfigure[High Data Uncertainty]{\label{fig:HighDataUnc}%
      \includegraphics[width=6cm]{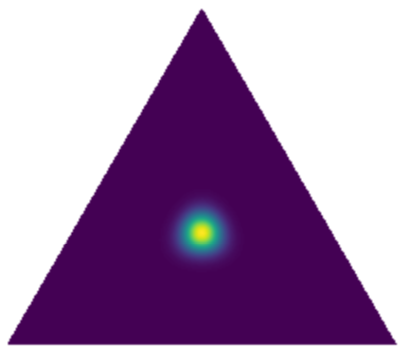}}%
    \qquad
    \subfigure[High Distributional Uncertainty]{\label{fig:HighDistUnc}%
      \includegraphics[width=6cm]{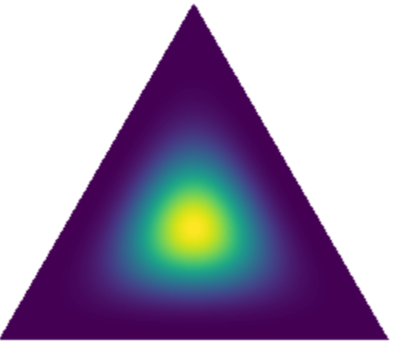}}
  }
\end{figure}

From Figure 1 it becomes clear that our method allows us to easily distinguish between Data and Distributional Uncertainty.
\newpage
\section{Additional Experiment}
\begin{figure}[htbp]
\centering
\includegraphics[width=14cm]{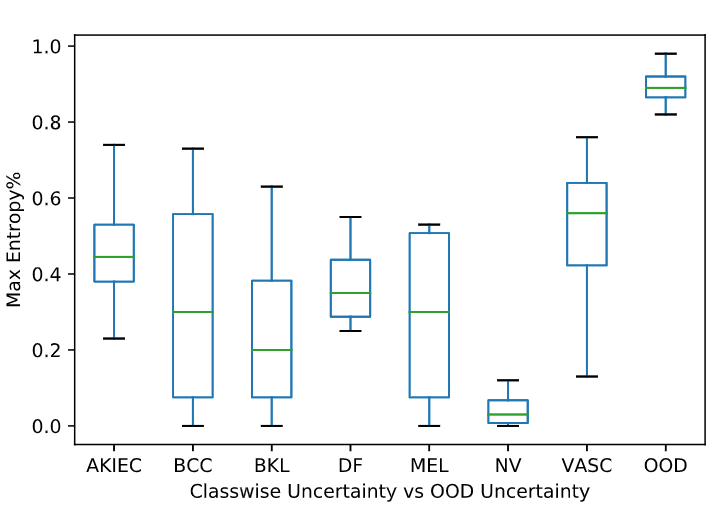}
\end{figure}
We observe our model is very confident on Nevi (NV) class, which is expected since it make majority of the dataset, this reinforces the importance of well balanced data for learning. We can also see our OOD samples can be distinctly separated from the in-class samples. The OOD sample used for training and testing are from different distributions. For simplicity we used Gaussian Distribution for training OOD samples and Uniform Distribution for testing OOD samples. Ideally more complex techniques should be used for generating OOD samples on the decision boundary
\citep{hafner2018noise}.
\end{document}